\pdfoutput=1

\documentclass[11pt]{article}

\usepackage[]{acl}

\usepackage{times}
\usepackage{latexsym}
\usepackage{amsmath,amsfonts,amssymb,amsthm}
\usepackage{comment}
\usepackage{booktabs}
\usepackage[ruled,noline,noend,lined]{algorithm2e}
\SetKwComment{Comment}{/* }{ */}
\usepackage{multirow}
\usepackage{array}

\SetAlCapNameFnt{\footnotesize}
\SetAlCapFnt{\footnotesize}

\usepackage{graphicx}

\usepackage[T1]{fontenc}

\usepackage[utf8]{inputenc}

\usepackage{microtype}
\newcommand\norm[1]{\left\lVert#1\right\rVert}

\title{On the Locality of Attention in Direct Speech Translation}

\author{Belen Alastruey\thanks{\hspace{1.3mm} Equal contribution.} , Javier Ferrando$^*$, Gerard I. Gállego \and Marta R. Costa-jussà \\
          TALP Research Center, Universitat Politècnica de Catalunya, Barcelona \\
         \texttt{\{belen.alastruey,javier.ferrando.monsonis} \\ \texttt{gerard.ion.gallego,marta.ruiz\}@upc.edu}}

\begin{document}
\maketitle
\begin{abstract}

Transformers have achieved state-of-the-art results across multiple NLP tasks. However, the self-attention mechanism complexity scales quadratically with the sequence length, creating an obstacle for tasks involving long sequences, like in the speech domain. In this paper, we discuss the usefulness of self-attention for Direct Speech Translation. First, we analyze the layer-wise token contributions in the self-attention of the encoder, unveiling local diagonal patterns. To prove that some attention weights are avoidable, we propose to substitute the standard self-attention with a local efficient one, setting the amount of context used based on the results of the analysis. With this approach, our model matches the baseline performance, and improves the efficiency by skipping the computation of those weights that standard attention discards.


\end{abstract}

\section{Introduction}
\label{sec:intro}

Recently, Transformer-based models have gained popularity and have revolutionized Natural Language Processing (NLP) \cite{NIPS2017_3f5ee243, devlin-etal-2019-bert, NEURIPS2020_1457c0d6}.
In the speech-to-text setting, the Transformer works with audio features like the mel-spectrogram \cite{Dong2018SpeechTransformerAN,gangi19_interspeech}. These features provide longer input sequences compared to their raw text counterparts. This can be a problem when regarding complexity, since the Transformer's attention matrix computational cost is $O(n^2)$, where $n$ is the sequence length. In speech, a common approach used to overcome this issue and reduce the input sequence length is to employ convolutional layers with stride before the Transformer encoder. 
However, even with the addition of convolutional layers, time and memory complexity is still an issue. 

An active area of research has investigated ways to make the Transformer more efficient in tasks involving long documents, that exhibit the same problem as speech tasks \cite{survey_efficient}. These models explore different techniques on how to avoid the computation of some attention weights, hence reducing the complexity of the self-attention layer. Some of these models, such as the Reformer \cite{reformer} or the Routing Transformer \cite{routing_transformer}, only compute attention weights on those queries and keys that are more related according to different clustering techniques.
\begin{figure}[!t]
    \centering
    \includegraphics[width=\linewidth]{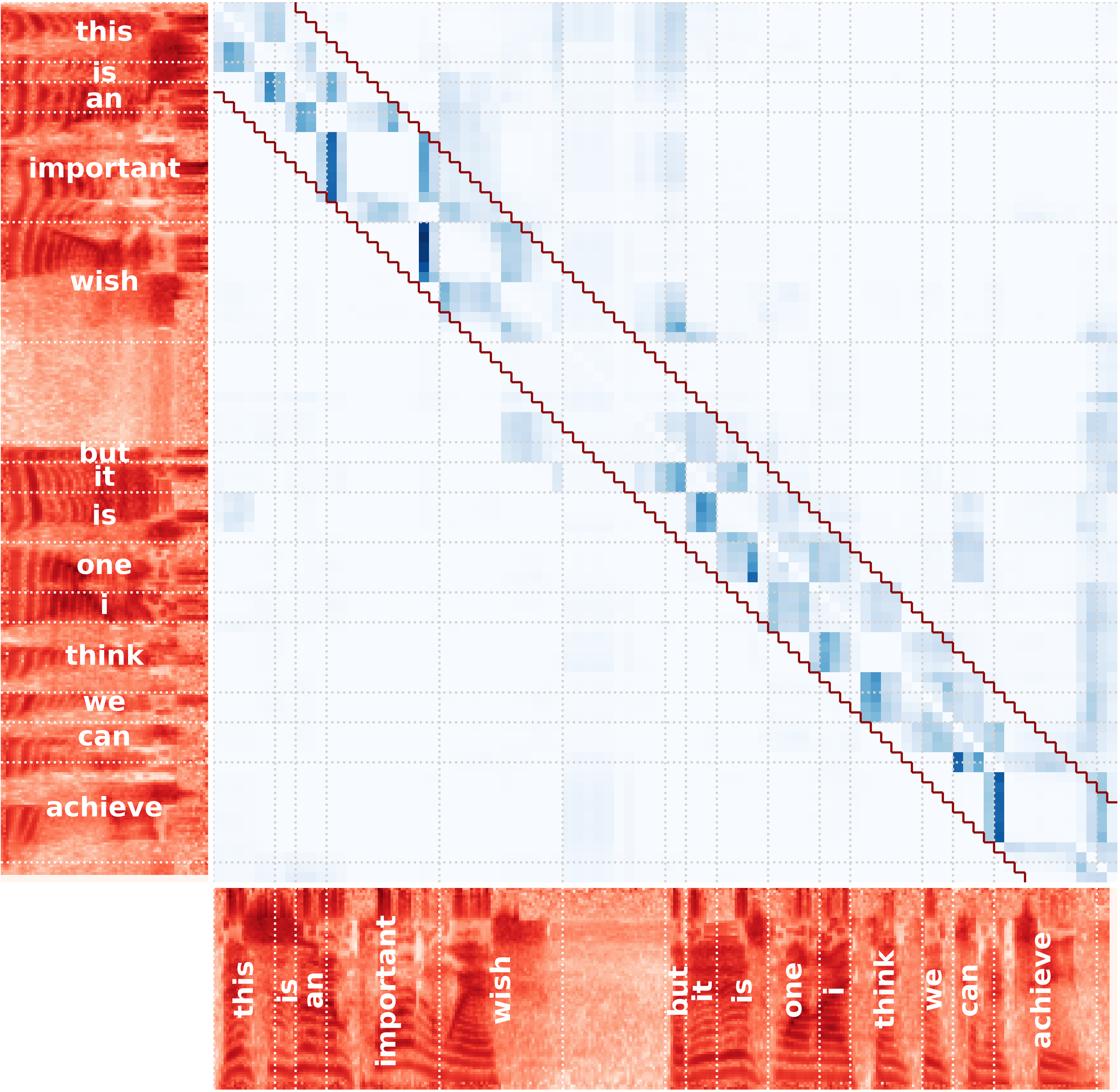}
    \caption{Spectrogram and contributions matrix\footnotemark in Layer 11, after training for En-De ST. Tokens attend locally, creating a diagonal pattern. Highlighted is shown our proposed adaptive local attention window.}
    \label{fig:spec_contrib}
\end{figure}
\footnotetext{The main diagonal, which accounts for 65\% of the total contributions, is hidden for visualization purposes.}
The authors of the Linformer \cite{linformer} state that the attention matrix is low-rank, so they project keys and values to reduce the size of the attention matrix. The Synthesizer \cite{synthesizer} directly avoids computing token-to-token interactions by learning synthetic attention weights. The Longformer \cite{longformer} and the Big Bird \cite{bigbird} modify the attention matrix with patterns such as local or random attentions. In this paper, we focus on local attention by using a sliding window centered on the diagonal of the attention matrix.


We build upon recent advances in the explainability of the Transformer to analyze the amount of context used by self-attention when dealing with speech features. Recent interpretability works have moved beyond raw attention weights as a measure of layer-wise input attributions and have integrated other modules in the self-attention, such as the norm of the vectors multiplying the attention weights \cite{kobayashi-etal-2020-attention}, the layer normalization, and the residual connection \cite{kobayashi-etal-2021-incorporating}. In the Automatic Speech Recognition (ASR) domain, the usefulness of the self-attention has been argued \cite{usefulness_self,shim2022understanding}, showing that its exposure to the full context might not be necessary, especially in the top layers.
We carry out this analysis for Direct Speech Translation (ST) systems, which are capable of translating between languages from speech to text with a single model. The encoder of these systems needs to jointly perform acoustic and semantic modeling, while in ASR the latter is not that relevant \cite{liu_2020}. To the best of our knowledge, this is the first work that uses interpretability methods to understand how the Transformer`s self-attention behaves in the Direct ST task.

In this work, we use the layer-wise contributions proposed by \citet{kobayashi-etal-2021-incorporating} to analyze the patterns of self-attention in Direct ST in En-De, En-Es and En-It tasks, unveiling their strong local nature. Consequently, using self-attention might not be entirely useful, but it is computationally costly. To verify this hypothesis, based on our analysis, we propose a new architecture designed to maximize the efficiency of the model while minimizing the information loss, and demonstrate no hinder in the model's performance in any of the three directions. We achieve this by substituting regular self-attention with local attention in those layers where the contributions are placed around the diagonal. Finally, we analyze the performance of the proposed model.

\section{Speech-to-text Transformer}
\label{sec:s2t_t}
Recent works have attempted to adapt the Transformer to speech tasks \cite{gangi19_interspeech, conformer}. In the Direct ST domain, a usual approach is adding two convolutional layers with a stride of 2 before the Transformer \cite{s2t_transformer}. By doing this, the sequence length is reduced to a fourth of the initial one. 
After the two convolutional layers, the speech-to-text Transformer (S2T Transformer) consists of a regular Transformer model, composed of 12 encoder layers and 6 decoder layers.

The main component of the Transformer is the multi-head attention mechanism, in particular, the self-attention is in charge of mixing contextual information. Given a sequence of token representations $\{\mathbf{x}_1,\cdots,\mathbf{x}_{N}\}$, each of the $H$ heads projects these vectors to queries $\mathbf{Q}^h \in \mathbb{R}^{N \times d_h}$, keys $\mathbf{K}^h \in \mathbb{R}^{N \times d_h}$ and values $\mathbf{V}^h \in \mathbb{R}^{N \times d_h}$, with head dimension $d_h=d/H$, where $d$ is the model embedding dimensionality. The self-attention attention (SA) computes:
\begin{equation}\label{eq:fsa}
\text{SA}(\mathbf{Q}^h, \mathbf{K}^h, \mathbf{V}^h) = \sum_{h}^{H} \mathbf{A}^{h}\mathbf{V}^h\mathbf{W}^{h}_{O} + \mathbf{b}_O
\end{equation}
Where $\mathbf{W}^{h}_{O} \in \mathbb{R}^{d_h \times d}$ and $\mathbf{b}_O \in \mathbb{R}^{d}$ are learnable parameters, and
\begin{equation}
\mathbf{A}^{h}=\text{softmax}\left(%
\frac{\mathbf{Q}^{h}(\mathbf{K}^{h})^T}{\sqrt{d_{h}}}\right)
\end{equation}
\paragraph{Training details.}
We reproduce the S2T Transformer training  with \textsc{Fairseq} \cite{ott-etal-2019-fairseq,s2t_transformer}. The training procedure consists of two phases. First, we pre-train the model in the ASR setting \cite{asrpretraining-st}. Then, we substitute the decoder with a randomly initialized one, and both are finally trained in the ST task (see Appendix \ref{apx:hyperparams} for more details on the hyperparameters). For the trainings, we use the MUST-C English-German, English-Spanish and English-Italian subsets \cite{mustc}.

\section{Model Analysis}\label{sec:s2t_int}
In this section, we present the analysis of the encoder self-attention in the S2T Transformer.

\paragraph{Interpretability method.} \citet{kobayashi-etal-2021-incorporating} propose an interpretability method that measures the impact of each layer input, i.e token representations ($\mathbf{x}_j$), to the output of the layer, considering also the layer normalization and the residual connection. They provide the formulation for the attention block of the original Transformer architecture, which has layer normalization on top of the self-attention module. In this work, we give an adaptation to the group of models that normalize before the multi-head attention (Pre-LN), such as the S2T Transformer. The complete chain of computations in the Pre-LN attention block can be reformulated as a simple expression of the layer inputs:
\begin{equation}
\hat{\mathbf{x}}_i= \sum_{j}^{N} \sum_{h}^{H} \mathbf{A}^{h}_{i,j}\text{LN}(\mathbf{x}_{j})\mathbf{W}_V^{h}\mathbf{W}^{h}_{O} + \mathbf{b}^O + \mathbf{x}_{i}
\end{equation}
We can now express the attention block output as a sum of transformed input vectors ($F_i(\mathbf{x}_j)$):
\begin{equation}\label{eq:pre_layer_transformed_vectors}
\hat{\mathbf{x}}_i = \sum_{j}^{N} F_i(\mathbf{x}_j) + \mathbf{b}^O
\end{equation}
Where $F_i(\mathbf{x}_j)$ is defined as:
\begin{equation*}
\resizebox{0.48\textwidth}{!}{$\displaystyle{
  F_i(\mathbf{x}_j)=\left\{
  \begin{array}{@{}ll@{}}
    \sum^H_h  \mathbf{A}_{i,j}^{h}\text{LN}(\mathbf{x}_{j})\mathbf{W}_V^{h}\mathbf{W}_O^{h} & \mbox{if}~ j \neq i \\
    \sum^H_h  \mathbf{A}_{i,j}^{h}\text{LN}(\mathbf{x}_{j})\mathbf{W}_V^{h}\mathbf{W}_O^{h} + \mathbf{x}_i & \mbox{if}~ j=i
  \end{array}\right.}$}
\end{equation*}
\citet{kobayashi-etal-2021-incorporating} measure the contribution $C_{i,j}$ of each input vector $\mathbf{x}_j$ to the layer output $\hat{\mathbf{x}}_i$ with the Euclidean norm of the transformed vector:
\begin{equation}
C_{i,j} = \norm{F_i(\mathbf{x}_j)}
\label{eq:c}
\end{equation}
 \begin{figure}[!t]
    \centering    \includegraphics[width=0.8\linewidth]{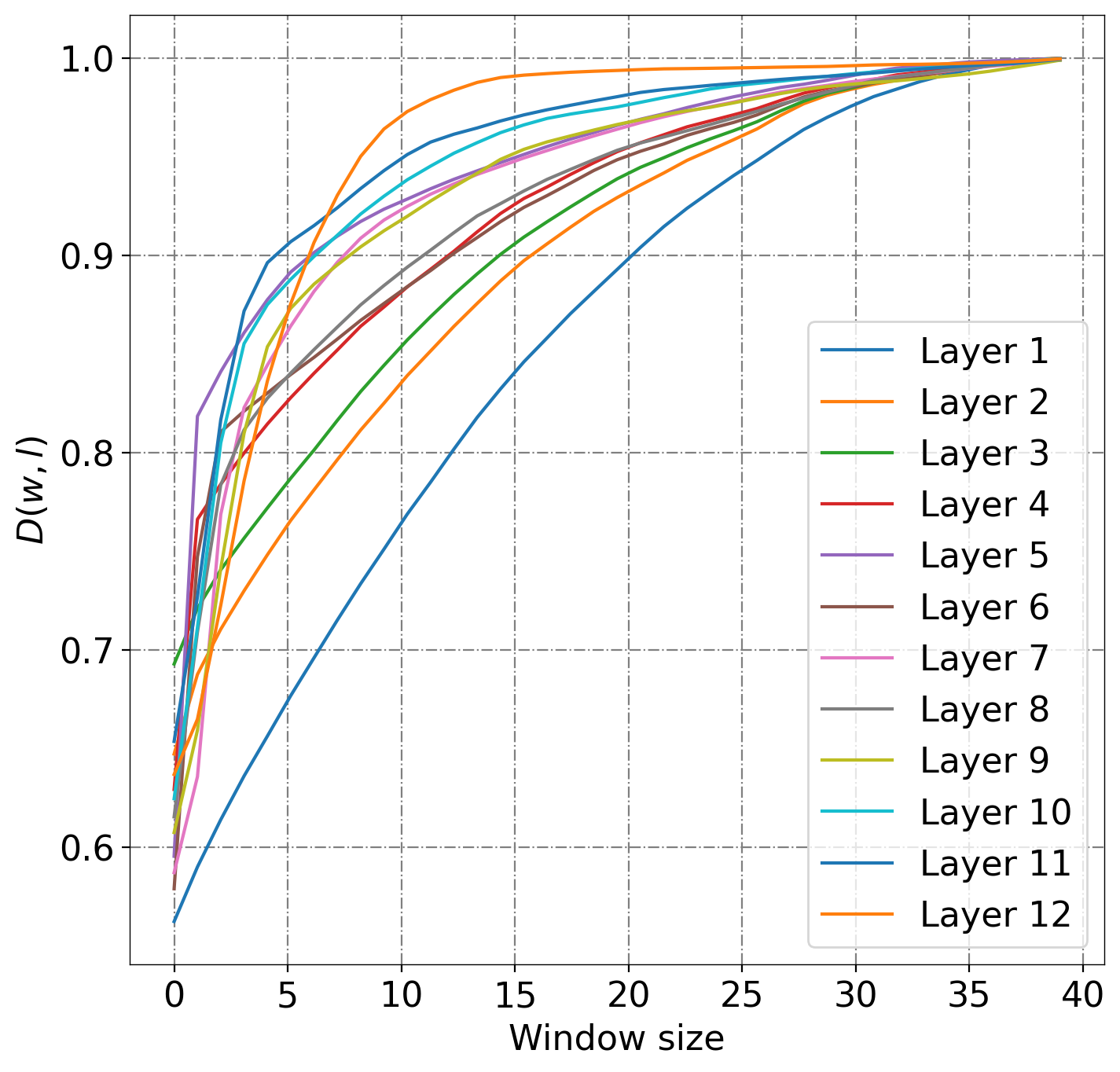}
    \caption{Contribution diagonality $D(w,l)$ after ST training, for a single En-De example. The greater the area under the curve (CCD), the higher the diagonality.}
    \label{fig:CAD}
\end{figure}

\begin{figure*}[!th]
\centering
\begin{minipage}[b]{\textwidth}
\begin{center}
\includegraphics[width=0.98\textwidth]{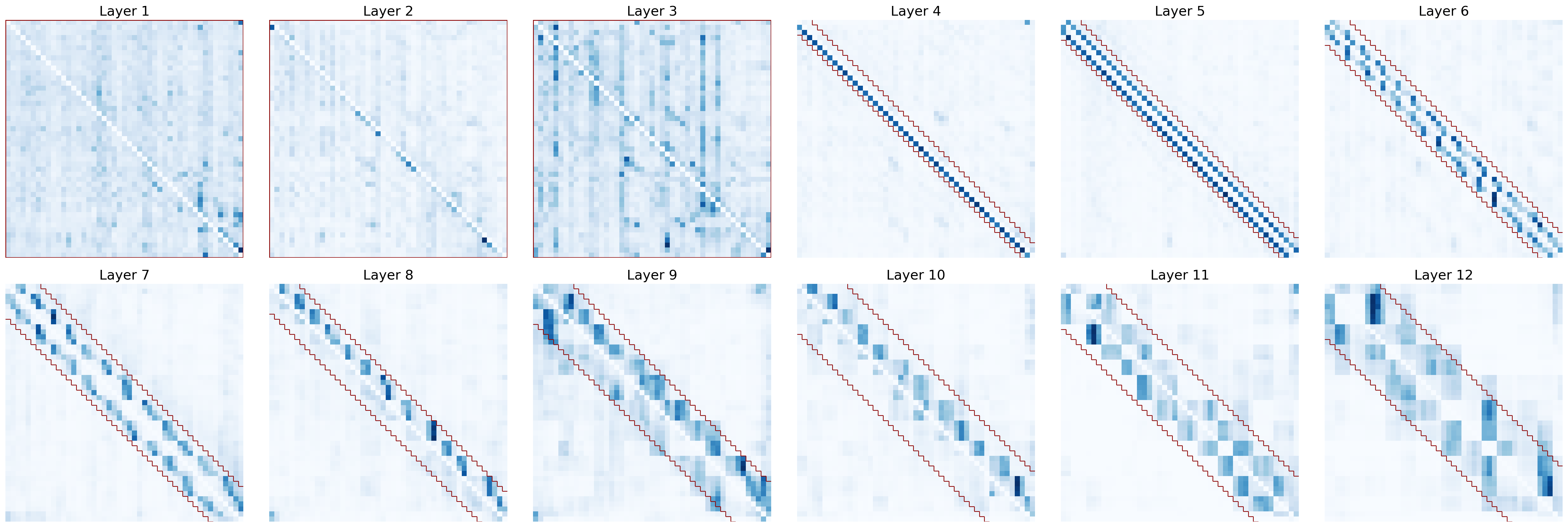}
\end{center}
\end{minipage}\hfill
\caption{Contribution matrices\footnotemark of encoder layers on a sample after training for En-De ST. Windows used in the efficient architecture highlighted.}
\label{fig:layers}
\end{figure*}

\paragraph{Layer-wise analysis.}
We analyze the contribution scores obtained with Eq. \ref{eq:c} from the encoder layers in both ASR (pre-training) and ST tasks. From the results shown in Figure \ref{fig:layers} (see also Appendix \ref{apx:layers_contributions}) we observe that most layers' contributions are dense around the diagonal. To measure the degree of diagonality in the contribution matrices at each layer $l$, we build upon the attention diagonality proposed by \citet{shim2022understanding}, originally defined with attention weights and proportions of the sequence length. We reformulate it with the obtained contributions, and token ranges $w$ (see Appendix \ref{apx:ccd_shim} for more details on the differences):
\begin{equation}
    D(w,l) = \frac{1}{N}\sum_{i}  \sum_{j} C_{i,j}^l
    \label{eq:d}
\end{equation}
where $j \in [\text{max}(1, i - \lfloor \frac{1}{2}w \rfloor), \text{min}(N, i + \lfloor \frac{1}{2}w \rfloor]$, $i \in [1,N]$. $D(w,l)$ computes the average of the contributions restricted by the diagonal window range $w$. In order to measure how fast the contribution density increases over the window length, we calculate the cumulative contribution diagonality (CCD), that corresponds to the area under the curve of the accumulated $D(w,l)$ within the range\footnote{Note that a window of size $w$ contains $\lfloor \frac{1}{2}w \rfloor$ tokens on each side of the main diagonal, so $w=1$ represents the main diagonal and $w=2N$ every possible diagonal. } $w \in [1,2N]$. That is, we approximate the integral of $D(l,d)$ along the distance $d$, but for the discrete variable $w$ (Figure \ref{fig:CAD}).

 \begin{figure}[!t]
    \centering
    \includegraphics[width=0.98\linewidth]{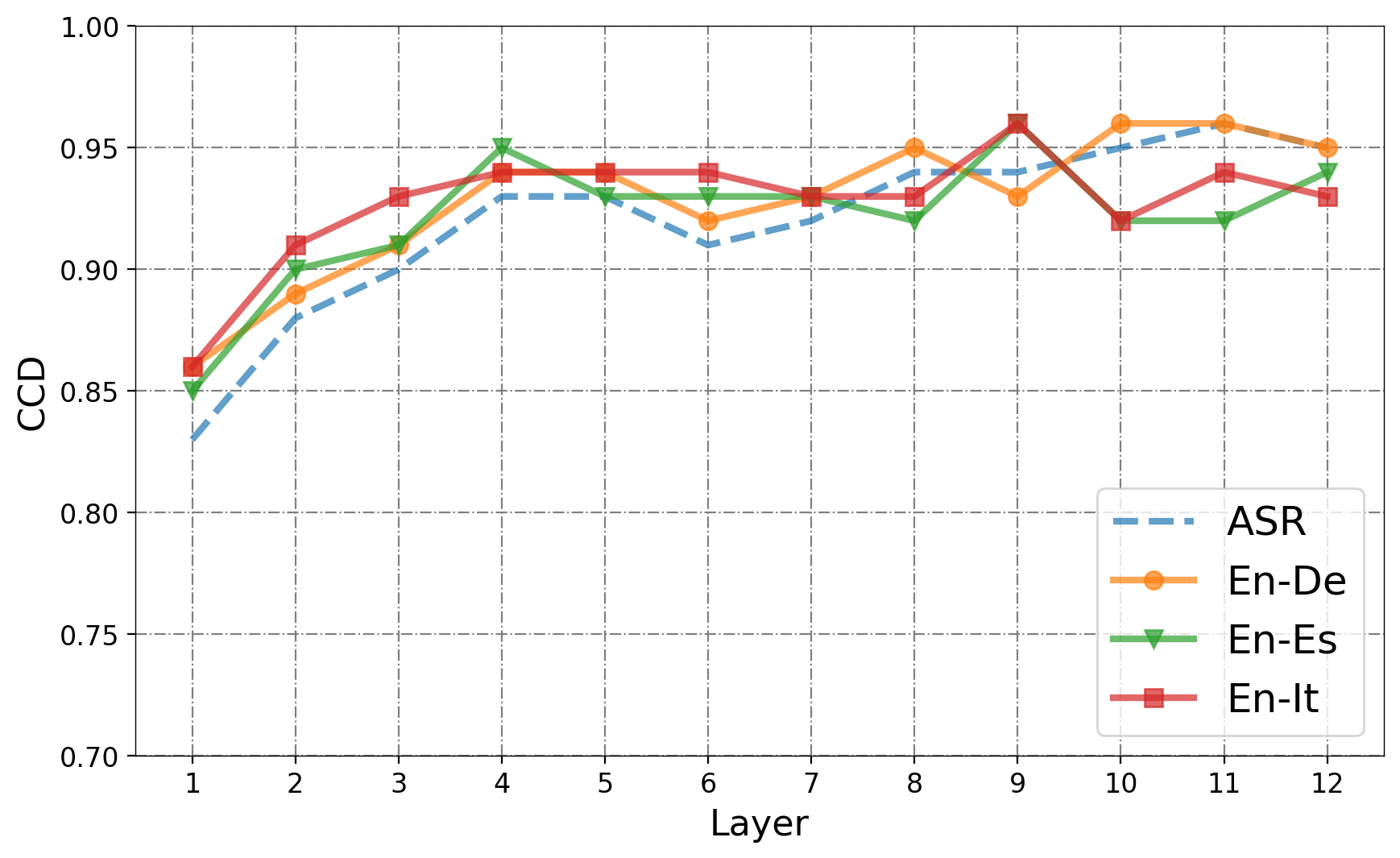}
    \caption{Average cumulative contribution diagonality (CCD) score across layers, over 100 samples. Results shown for models trained in ASR (dashed line) and ST (solid line).}
    \label{fig:diag}
\end{figure}

In Figure \ref{fig:diag} we show the CCD results for ASR and ST across layers, where we can observe a strong diagonal pattern. We can see that, surprisingly, CCD is very similar in both tasks. This contradicts the belief that, because of the need for deeper semantic processing when translating, ST needs more context than ASR. Furthermore, we see different behaviors along the encoder, and a trend towards uniformly distributed contributions in the first layers. 

Moreover, in Figure \ref{fig:layers}, we can see differences between those layers that show local patterns. Layers 4, 5 and 6 attend to close context. Instead, those layers at the end of the encoder, as 11 or 12, need larger context, and we can see how the contributions create patterns corresponding to words in the spectrogram, enabling us to see interactions between them (Figure \ref{fig:spec_contrib}). Additionally, we see that contribution matrices reveal silences in the speech sequence. However, we believe that further research is needed to fully understand the meaning of these patterns.

\footnotetext{The main diagonal, which accounts for around 65\% of the total contributions in each layer, is hidden for visualization purposes.}

\section{Efficient Speech-to-text Transformer}
\label{sec:eff_s2t_t}
From the previous analysis, we hypothesize that suitable local attention patterns may potentially avoid the computation of unused attention scores. Note that if a token does not contribute to the output of a layer, its attention score can be canceled. Our objective is to maximize the efficiency of the model while minimizing the performance drop.

\begin{algorithm}[!h]
\DontPrintSemicolon
\footnotesize
\caption{\small Window size selection}\label{alg:cap}
\KwIn{\;$C^l$: contribution matrix, $N$: number of tokens,\\
$t$: min diagonal contribution threshold}
\KwOut{\\$w^l$: optimal window size}\;
\SetInd{0.65em}{0.1em}
 $counter \gets 0$\;
 $w^l \gets 0$\;
\While {$counter < (N/10)$}{
    \For{$i \gets 0,  N$}{
        \uIf{$\text{mean}(C^l_{diag}[i]) > t \; \text{\textbf{or}} \; \text{mean}(C^l_{diag}[-i]) > t$\;}{
             $w^l \gets 2i +1$\;
             $counter \gets 0$\;}
        {\Else{
             $counter \gets counter + 1$\;}
        }
    }
}
\end{algorithm}

\paragraph{Window size selection.} CCD could serve as a starting point to obtain optimal window lengths. However, it requires predefining the amount of total contribution required inside the window, which makes it fragile to properly detect local patterns. On one hand, it can be too sensitive to a strong main diagonal. On the other, it may overestimate random distant contributions. We propose an alternative based on the average contribution of every sub/superdiagonal (Alg. \ref{alg:cap}). Starting from the main diagonal $C_{diag}[0]$, it keeps adding tokens to the window length until it finds $N \cdot 10\%$ consecutive sub/superdiagonals below the $t$ threshold\footnote{Hyperparameter that defines the minimum average value of the sub/superdiagonals to be considered. We choose 0.01 after empirical study.}. We repeat this procedure with a random set of 400 sentences, and we compute each layer's window length mean ($\mu^l$) and standard deviation ($\sigma^l$). To ensure that most significant contributions are considered, we define as the optimal window size ($w^l$) the result of the operation\footnote{If $w^l$ is even, we set $w^l = w^l + 1$ so that it is odd and hence the window can be centered around the diagonal.} $w^l = \lceil \mu^l + \sigma^l \rceil$. The results obtained are similar in every language pair (Table \ref{tab:windows} and Appendix \ref{apx:languges}). In Figure \ref{fig:layers}, we can see how $w$ contains most of the relevant contributions for En-De ST (see more examples in Appendix \ref{apx:layers_contributions}).

\begin{table}[!t]
\small
\centering
\begin{tabular}{lccc}
\toprule
\textbf{Layer} & $\boldsymbol{\mu}\pm \boldsymbol{\sigma}$ & $\boldsymbol{w}$ & \textbf{$CL$}\\
\midrule
\textbf{1 *} & $  \mathbin{\:}\ 3.41\pm 13.15$ & \textit{17} & \textit{0.35} $\pm$ \textit{0.07} \\
\textbf{2 *} & $  \mathbin{\:}\ 1.18 \pm 3.45  \mathbin{\:}\ $ & \textit{5} & \textit{0.32} $\pm$ \textit{0.04}\\
\textbf{3 *} & $  \mathbin{\:}\ 0.51 \pm 1.56 \mathbin{\:}\ $ & \textit{3} & \textit{0.30} $\pm$ \textit{0.04} \\ 
\textbf{4} & $  \mathbin{\:}\ 2.25 \pm 1.30  \mathbin{\:}\ $ & 5 & $0.23 \pm 0.04$ \\ 
\textbf{5} & $ \mathbin{\:}\ 4.03 \pm 0.28  \mathbin{\:}\ $ & 5 & $0.17 \pm 0.03$ \\
\textbf{6} & $ \mathbin{\:}\ 7.03 \pm 1.03  \mathbin{\:}\ $ & 9 & $0.23 \pm 0.04$ \\ 
\textbf{7} & $11.37 \pm 1.13  \mathbin{\:}\ $ & 13 & $0.18 \pm 0.04$ \\
\textbf{8} & $\mathbin{\:}\ 7.94 \pm 1.16 \mathbin{\:}\ $ & 11 & $0.18 \pm 0.04$ \\
\textbf{9} & $12.56 \pm 1.85 \mathbin{\:}\ $ & 15 & $0.19 \pm 0.05$ \\
\textbf{10} & $16.47 \pm 2.40 \mathbin{\:}\ $ & 19 & $0.13 \pm 0.05$ \\
\textbf{11} & $13.28 \pm 1.90 \mathbin{\:}\ $ & 17 & $0.13 \pm 0.04$ \\
\textbf{12} & $16.28 \pm 3.86 \mathbin{\:}\ $ & 21 & $0.16 \pm 0.05$ \\
\bottomrule
\end{tabular}
\caption{Optimal window size study in En-De ST. (*) For the first three layers, we use standard self-attention.}
\label{tab:windows}
\end{table}
\paragraph{Contribution loss.} We can now calculate the percentage of the total contributions that are left outside the window. This allows us to discover the amount of contribution that is lost because of the use of local attention. To do so, we employ Eq. \ref{eq:d}, but since we are interested in the contributions outside each window $w^l$, we define $CL(l,w^l) = 1 - D(l,w^l)$.

\paragraph{Proposed architecture.} From the previous results, we see that the first three layers are the ones with the weakest local pattern (See Figure \ref{fig:layers}). In these layers, $CL(l, w^l)$ is large, and CCD (Figure \ref{fig:CAD}) shows smaller areas. For these reasons, we believe that using the entire self-attention in the first three layers is necessary. In the following layers, we use local attention with window size $w^l$.  Our proposed architecture is an efficient adaptation of the S2T Transformer, and therefore it is exactly equal with exception of the self-attention layers (detailed architectures in Section \ref{sec:s2t_t} and Appendix \ref{apx:arch_details}).

\paragraph{Experiments.}
Finally, we train our model under the same specifications and dataset as the baseline (see Section \ref{sec:s2t_t} for details on the dataset, and Appendix \ref{apx:hyperparams} for the training hyperparameters).


\begin{table}[t]
\small
\centering
\resizebox{\columnwidth}{!}{
\begin{tabular}{lccc}
\toprule
 & \textbf{En-De} & \textbf{En-Es} & \textbf{En-It}\\
\midrule
\textit{Baseline} & $ 22.53 \pm 0.15 $ & $ 27.49 \pm 0.22 $ & $ 22.98 \pm 0.15 $ \\
\textit{Ours} & $ 22.49 \pm 0.11 $ & $ 27.46 \pm 0.12 $ & $ 22.97 \pm 0.27 $\\
\bottomrule
\end{tabular}
}
\caption{BLEU obtained on the Speech Translation task (mean $\pm$ std after training with 5 different seeds).}
\label{tab:results}
\end{table}


As we see in Table \ref{tab:results}, our model matches the performance of the S2T Transformer in every analyzed language pair. However, we achieve it while reducing the complexity in most layers from $O(n^2)$ to $O(n \cdot w^l)$. This difference can be highly significant, considering the usual length of speech sequences and the size of the windows used.  In particular, $w^l$ goes from 5 to 25 tokens between the different languages. However, the average length of an input sequence in studied splits of the MUST-C dataset after the two convolutional layers used in the S2T Transformer, is 166 tokens, even reaching a maximum of 1052.


\section{Conclusions}
\label{sec:conclusion}
Transformer-based models are the current state-of-the-art in many different fields. However, the quadratic complexity of the self-attention module usually hinders the usefulness of the model in real-life applications. This problem worsens when working with long sequences, as is the case with speech. In this paper, we have questioned the need of computing all attention weights in ST. We have analyzed the contribution matrices, and we have seen that, in many layers, the relevant scores are placed in a diagonal pattern. Therefore, we have hypothesized that these weights do not need to be calculated. To verify our hypothesis, we have trained a model that substitutes regular self-attention with local attention, with a suitable window size for each layer. We have seen that as we expected, the results are almost equal to the ones obtained with the baseline model, but the complexity has been lowered significantly.

Regarding interpretability, we have found how the Transformer establishes connections between words in speech sequences. Furthermore, we have seen that, in contrast to what was expected, diagonality scores are similar in both ST and ASR tasks, meaning that they use the same amount of context. 

\section{Acknowledgments}
This work was partially funded by the project ADAVOICE, PID2019-107579RB-I00 / AEI / 10.13039/501100011033, and the UPC INIREC scholarship nº3522. We would like to thank Ioannis Tsiamas and Carlos Escolano for their support and advice, and the anonymous reviewers for their useful comments.

\section{Ethical Considerations}
This work analyzes the inner workings of a particular architecture in Direct Speech Translation. Based on the analysis, we propose a more efficient model, that maintains the baseline performance. Our proposed solution can help reduce the ecological footprint of Speech Translation systems based on the Transformer architecture. We believe this work has no direct negative social influences. However, we should underline that the dataset used in this paper consists of high-resource languages such as English, German, Spanish, and Italian. Although the interpretability method does not depend on specific languages, there may be differences in the degree of efficiency that can be achieved when experimenting with other languages.

\bibliography{anthology,custom}

\begin{thebibliography}{22}
\expandafter\ifx\csname natexlab\endcsname\relax\def\natexlab#1{#1}\fi

\bibitem[{Beltagy et~al.(2020)Beltagy, Peters, and Cohan}]{longformer}
Iz~Beltagy, Matthew~E. Peters, and Arman Cohan. 2020.
\newblock Longformer: The long-document transformer.
\newblock \emph{ArXiv}, abs/2004.05150.

\bibitem[{Brown et~al.(2020)Brown, Mann, Ryder, Subbiah, Kaplan, Dhariwal,
  Neelakantan, Shyam, Sastry, Askell, Agarwal, Herbert-Voss, Krueger, Henighan,
  Child, Ramesh, Ziegler, Wu, Winter, Hesse, Chen, Sigler, Litwin, Gray, Chess,
  Clark, Berner, McCandlish, Radford, Sutskever, and
  Amodei}]{NEURIPS2020_1457c0d6}
Tom Brown, Benjamin Mann, Nick Ryder, Melanie Subbiah, Jared~D Kaplan, Prafulla
  Dhariwal, Arvind Neelakantan, Pranav Shyam, Girish Sastry, Amanda Askell,
  Sandhini Agarwal, Ariel Herbert-Voss, Gretchen Krueger, Tom Henighan, Rewon
  Child, Aditya Ramesh, Daniel Ziegler, Jeffrey Wu, Clemens Winter, Chris
  Hesse, Mark Chen, Eric Sigler, Mateusz Litwin, Scott Gray, Benjamin Chess,
  Jack Clark, Christopher Berner, Sam McCandlish, Alec Radford, Ilya Sutskever,
  and Dario Amodei. 2020.
\newblock \href
  {https://proceedings.neurips.cc/paper/2020/file/1457c0d6bfcb4967418bfb8ac142f64a-Paper.pdf}
  {Language models are few-shot learners}.
\newblock In \emph{Advances in Neural Information Processing Systems},
  volume~33, pages 1877--1901. Curran Associates, Inc.

\bibitem[{Bérard et~al.(2018)Bérard, Besacier, Kocabiyikoglu, and
  Pietquin}]{asrpretraining-st}
Alexandre Bérard, Laurent Besacier, Ali~Can Kocabiyikoglu, and Olivier
  Pietquin. 2018.
\newblock \href {https://doi.org/10.1109/ICASSP.2018.8461690} {End-to-end
  automatic speech translation of audiobooks}.
\newblock In \emph{2018 IEEE International Conference on Acoustics, Speech and
  Signal Processing (ICASSP)}, pages 6224--6228.

\bibitem[{Cattoni et~al.(2021)Cattoni, {Di Gangi}, Bentivogli, Negri, and
  Turchi}]{mustc}
Roldano Cattoni, Mattia~Antonino {Di Gangi}, Luisa Bentivogli, Matteo Negri,
  and Marco Turchi. 2021.
\newblock \href {https://doi.org/https://doi.org/10.1016/j.csl.2020.101155}
  {Must-c: A multilingual corpus for end-to-end speech translation}.
\newblock \emph{Computer Speech \& Language}, 66:101155.

\bibitem[{Devlin et~al.(2019)Devlin, Chang, Lee, and
  Toutanova}]{devlin-etal-2019-bert}
Jacob Devlin, Ming-Wei Chang, Kenton Lee, and Kristina Toutanova. 2019.
\newblock \href {https://doi.org/10.18653/v1/N19-1423} {{BERT}: Pre-training of
  deep bidirectional transformers for language understanding}.
\newblock In \emph{Proceedings of the 2019 Conference of the North {A}merican
  Chapter of the Association for Computational Linguistics: Human Language
  Technologies, Volume 1 (Long and Short Papers)}, pages 4171--4186,
  Minneapolis, Minnesota. Association for Computational Linguistics.

\bibitem[{{Di Gangi} et~al.(2019){Di Gangi}, Negri, and
  Turchi}]{gangi19_interspeech}
Mattia~Antonino {Di Gangi}, Matteo Negri, and Marco Turchi. 2019.
\newblock \href {https://doi.org/10.21437/Interspeech.2019-3045} {{Adapting
  Transformer to End-to-End Spoken Language Translation}}.
\newblock In \emph{Proc. Interspeech 2019}, pages 1133--1137.

\bibitem[{Dong et~al.(2018)Dong, Xu, and Xu}]{Dong2018SpeechTransformerAN}
Linhao Dong, Shuang Xu, and Bo~Xu. 2018.
\newblock Speech-transformer: A no-recurrence sequence-to-sequence model for
  speech recognition.
\newblock \emph{2018 IEEE International Conference on Acoustics, Speech and
  Signal Processing (ICASSP)}, pages 5884--5888.

\bibitem[{Gulati et~al.(2020)Gulati, Qin, Chiu, Parmar, Zhang, Yu, Han, Wang,
  Zhang, Wu, and Pang}]{conformer}
Anmol Gulati, James Qin, Chung-Cheng Chiu, Niki Parmar, Yu~Zhang, Jiahui Yu,
  Wei Han, Shibo Wang, Zhengdong Zhang, Yonghui Wu, and Ruoming Pang. 2020.
\newblock \href {https://doi.org/10.21437/Interspeech.2020-3015} {{Conformer:
  Convolution-augmented Transformer for Speech Recognition}}.
\newblock In \emph{Proc. Interspeech 2020}, pages 5036--5040.

\bibitem[{Kitaev et~al.(2020)Kitaev, Kaiser, and Levskaya}]{reformer}
Nikita Kitaev, {\L}ukasz Kaiser, and Anselm Levskaya. 2020.
\newblock Reformer: The efficient transformer.
\newblock In \emph{International Conference on Learning Representations}.

\bibitem[{Kobayashi et~al.(2020)Kobayashi, Kuribayashi, Yokoi, and
  Inui}]{kobayashi-etal-2020-attention}
Goro Kobayashi, Tatsuki Kuribayashi, Sho Yokoi, and Kentaro Inui. 2020.
\newblock \href {https://doi.org/10.18653/v1/2020.emnlp-main.574} {Attention is
  not only a weight: Analyzing transformers with vector norms}.
\newblock In \emph{Proceedings of the 2020 Conference on Empirical Methods in
  Natural Language Processing (EMNLP)}, pages 7057--7075, Online. Association
  for Computational Linguistics.

\bibitem[{Kobayashi et~al.(2021)Kobayashi, Kuribayashi, Yokoi, and
  Inui}]{kobayashi-etal-2021-incorporating}
Goro Kobayashi, Tatsuki Kuribayashi, Sho Yokoi, and Kentaro Inui. 2021.
\newblock \href {https://doi.org/10.18653/v1/2021.emnlp-main.373}
  {{I}ncorporating {R}esidual and {N}ormalization {L}ayers into {A}nalysis of
  {M}asked {L}anguage {M}odels}.
\newblock In \emph{Proceedings of the 2021 Conference on Empirical Methods in
  Natural Language Processing}, pages 4547--4568, Online and Punta Cana,
  Dominican Republic. Association for Computational Linguistics.

\bibitem[{Liu et~al.(2020)Liu, Zhu, Zhang, and Zong}]{liu_2020}
Yuchen Liu, Junnan Zhu, Jiajun Zhang, and Chengqing Zong. 2020.
\newblock Bridging the modality gap for speech-to-text translation.
\newblock \emph{ArXiv}, abs/2010.14920.

\bibitem[{Ott et~al.(2019)Ott, Edunov, Baevski, Fan, Gross, Ng, Grangier, and
  Auli}]{ott-etal-2019-fairseq}
Myle Ott, Sergey Edunov, Alexei Baevski, Angela Fan, Sam Gross, Nathan Ng,
  David Grangier, and Michael Auli. 2019.
\newblock \href {https://doi.org/10.18653/v1/N19-4009} {fairseq: A fast,
  extensible toolkit for sequence modeling}.
\newblock In \emph{Proceedings of the 2019 Conference of the North {A}merican
  Chapter of the Association for Computational Linguistics (Demonstrations)},
  pages 48--53, Minneapolis, Minnesota. Association for Computational
  Linguistics.

\bibitem[{Roy et~al.(2021)Roy, Saffar, Vaswani, and
  Grangier}]{routing_transformer}
Aurko Roy, Mohammad Saffar, Ashish Vaswani, and David Grangier. 2021.
\newblock Efficient content-based sparse attention with routing transformers.
\newblock \emph{Transactions of the Association for Computational Linguistics},
  9:53--68.

\bibitem[{Shim et~al.(2022)Shim, Choi, and Sung}]{shim2022understanding}
Kyuhong Shim, Jungwook Choi, and Wonyong Sung. 2022.
\newblock \href {https://openreview.net/forum?id=AvcfxqRy4Y} {Understanding the
  role of self attention for efficient speech recognition}.
\newblock In \emph{International Conference on Learning Representations}.

\bibitem[{Tay et~al.(2021)Tay, Bahri, Metzler, Juan, Zhao, and
  Zheng}]{synthesizer}
Yi~Tay, Dara Bahri, Donald Metzler, Da-Cheng Juan, Zhe Zhao, and Che Zheng.
  2021.
\newblock Synthesizer: Rethinking self-attention in transformer models.
\newblock In \emph{International Conference on Machine Learning}. PMLR.

\bibitem[{Tay et~al.(2020)Tay, Dehghani, Bahri, and Metzler}]{survey_efficient}
Yi~Tay, Mostafa Dehghani, Dara Bahri, and Donald Metzler. 2020.
\newblock Efficient transformers: A survey.
\newblock \emph{ArXiv}, abs/2009.06732.

\bibitem[{Vaswani et~al.(2017)Vaswani, Shazeer, Parmar, Uszkoreit, Jones,
  Gomez, Kaiser, and Polosukhin}]{NIPS2017_3f5ee243}
Ashish Vaswani, Noam Shazeer, Niki Parmar, Jakob Uszkoreit, Llion Jones,
  Aidan~N Gomez, \L~ukasz Kaiser, and Illia Polosukhin. 2017.
\newblock \href
  {https://proceedings.neurips.cc/paper/2017/file/3f5ee243547dee91fbd053c1c4a845aa-Paper.pdf}
  {Attention is all you need}.
\newblock In \emph{Advances in Neural Information Processing Systems},
  volume~30. Curran Associates, Inc.

\bibitem[{Wang et~al.(2020{\natexlab{a}})Wang, Tang, Ma, Wu, Okhonko, and
  Pino}]{s2t_transformer}
Changhan Wang, Yun Tang, Xutai Ma, Anne Wu, Dmytro Okhonko, and Juan Pino.
  2020{\natexlab{a}}.
\newblock \href {https://aclanthology.org/2020.aacl-demo.6} {Fairseq {S}2{T}:
  Fast speech-to-text modeling with fairseq}.
\newblock In \emph{Proceedings of the 1st Conference of the Asia-Pacific
  Chapter of the Association for Computational Linguistics and the 10th
  International Joint Conference on Natural Language Processing: System
  Demonstrations}, pages 33--39, Suzhou, China. Association for Computational
  Linguistics.

\bibitem[{Wang et~al.(2020{\natexlab{b}})Wang, Li, Khabsa, Fang, and
  Ma}]{linformer}
Sinong Wang, Belinda~Z. Li, Madian Khabsa, Han Fang, and Hao Ma.
  2020{\natexlab{b}}.
\newblock Linformer: Self-attention with linear complexity.
\newblock \emph{ArXiv}, abs/2006.04768.

\bibitem[{Zaheer et~al.(2020)Zaheer, Guruganesh, Dubey, Ainslie, Alberti,
  Ontanon, Pham, Ravula, Wang, Yang, and Ahmed}]{bigbird}
Manzil Zaheer, Guru Guruganesh, Kumar~Avinava Dubey, Joshua Ainslie, Chris
  Alberti, Santiago Ontanon, Philip Pham, Anirudh Ravula, Qifan Wang, Li~Yang,
  and Amr Ahmed. 2020.
\newblock \href
  {https://proceedings.neurips.cc/paper/2020/file/c8512d142a2d849725f31a9a7a361ab9-Paper.pdf}
  {Big bird: Transformers for longer sequences}.
\newblock In \emph{Advances in Neural Information Processing Systems},
  volume~33, pages 17283--17297. Curran Associates, Inc.

\bibitem[{Zhang et~al.(2021)Zhang, Loweimi, Bell, and Renals}]{usefulness_self}
Shucong Zhang, Erfan Loweimi, Peter Bell, and Steve Renals. 2021.
\newblock \href {https://doi.org/10.1109/SLT48900.2021.9383521} {On the
  usefulness of self-attention for automatic speech recognition with
  transformers}.
\newblock In \emph{2021 IEEE Spoken Language Technology Workshop (SLT)}, pages
  89--96.

\end{thebibliography}
\bibliographystyle{acl_natbib}
\newpage
\appendix

\section{Cumulative Attention Diagonality (CAD)}
\label{apx:ccd_shim}
\citet{shim2022understanding} propose the cumulative attention diagonality (CAD) as the integral of the attention diagonality $D(r,l)$ along the variable $r$, which defines the window length as a proportion of the sequence length:

\begin{equation*}
    CAD^l =  \int_{r=0}^{r=1} D(r,l)\;dr
\end{equation*}
where $D(r,l)$ is defined over the attention weight matrix $A^l$:
\begin{equation*}
    D(r,l) = \frac{1}{N}\sum_{i = 1}^{N}   \sum_{\substack{ j = \text{max}(1, \\ i - r(N-1) }}^{\substack{\text{min}(N, \\ i + r(N-1))}} A_{i,j}^l 
\end{equation*}
To approximate the result of the integral, \citet{shim2022understanding} use the Trapezoidal Rule with the discretized variable $\hat{r} \approx r$.
\begin{equation*}
    \int_{r=0}^{r=1} D(r,l)\;dr \approx \sum_{\hat{r}=0}^{\hat{r}=1} \frac{D(\hat{r},l) + D(\hat{r}+1,l)}{2}
\end{equation*}
For each step in the summation, the window range around the diagonal increases $2r(N-1)$, which may lead to different increments based on the sentence length. For instance, for a sentence with $N=11$, in a 0.1 increase of $\hat{r}$, the window size range increases by $2$. However, with $N=101$ we get an increment of $20$. For this reason, we redefine the diagonality measures with token-wise increments.
 
\section{Architecture Details}
\label{apx:arch_details}
Both our efficient model and the S2T Transformer \cite{s2t_transformer} share the same architecture, with the exception of the self-attention modules.
The models consist 12 encoder layers and 6 decoder layers with sinusoidal positional encodings. In the encoder and decoder we use 4 attention heads, an embedding dimension of 256, and of 2048 in the FFN layers. We use a dropout probability of 0.1 in both the attention weights and FFN activations. We use ReLU as the activation function.

Regarding the convolutional layer applied to reduce sequence length, it consists of a 1D convolutional layer, with a kernel of size 5, a stride of 2, and with the same number of output channels than input channels.

\section{Training Hyperparameters}
\label{apx:hyperparams}
To ensure a reliable comparison, we performed all ASR and ST experiments under the same conditions and hyperparameters. In ASR training we fixed a maximum of 40000 tokens per batch. We used Adam optimizer and a learning rate of $1 \cdot 10^{-3}$ with an inverse square root scheduler. We applied a warm-up for the first 10000 updates. We clipped the gradient to 10 to avoid exploding gradients. We used label smoothed cross-entropy as a loss function, with a smoothing factor of 0.1. We used an update frequency of 8 on a single GPU. We set a maximum of 50000 updates for every training.
In ST training, we use the same hyperparameters as for ASR, but we use a learning rate of $2 \cdot 10^{-3}$.
We conducted the training of all our experiments using NVIDIA GeForce RTX 2080 Ti GPU.

\section{Optimal window analysis in En-Es and En-It ST}
\label{apx:languges}

\begin{table}[!h]
\small
\centering
\begin{tabular}{lccc}
\toprule
\textbf{Layer} & $\boldsymbol{\mu}\pm \boldsymbol{\sigma}$ & $\boldsymbol{w}$ & \textbf{$CL$}\\
\midrule
\textbf{1 *} & $4.68 \pm 14.77$ & \textit{21} & \textit{0.39} $\pm$ \textit{0.08} \\
\textbf{2 *} & $ 3.21 \pm 6.17$ & \textit{11} & \textit{0.29} $\pm$ \textit{0.04}\\
\textbf{3 *} & $ 0.99 \pm 3.6 $ & \textit{5} & \textit{0.28} $\pm$ \textit{0.04} \\ 
\textbf{4} & $  2.58\pm1.96 $ & 5 & $ 0.2\pm0.02 $ \\ 
\textbf{5} & $ 4.52 \pm 2.38$ & 7 & $ 0.24\pm0.04 $ \\
\textbf{6} & $ 15.88\pm2.92 $ & 19 & $ 0.21\pm0.06 $ \\
\textbf{7} & $ 11.32\pm1.91 $ & 15 & $0.16 \pm 0.03$ \\
\textbf{8} & $ 9.52 \pm 2.5$& 13 & $ 0.22\pm0.05 $ \\
\textbf{9} & $ 14.96\pm1.78 $ & 17 & $0.07\pm0.05 $ \\
\textbf{10} & $ 15.94\pm 3.0$ & 19 & $ 0.19 \pm0.05 $ \\
\textbf{11} & $13.83 \pm 3.66$ & 19 & $ 0.21\pm0.05 $ \\
\textbf{12} & $ 20.38 \pm 3.42$ & 25 & $0.1 \pm 0.05$ \\
\bottomrule
\end{tabular}
\caption{Optimal window size study in En-Es ST. (*) For the first three layers, we use standard self-attention.}
\label{tab:windows_es}
\end{table}

\begin{table}[!h]
\small
\centering
\begin{tabular}{lccc}
\toprule
\textbf{Layer} & $\boldsymbol{\mu}\pm \boldsymbol{\sigma}$ & $\boldsymbol{w}$ & \textbf{$CL$}\\
\midrule
\textbf{1 *} & $ 6.16\pm 17.57$ & \textit{25} & \textit{0.34} $ \pm $ \textit{0.09} \\
\textbf{2 *} & $ 2.56\pm 7.47$ & \textit{11} & \textit{0.29} $ \pm $ \textit{0.05}\\
\textbf{3 *} & $ 2.44 \pm 2.84$ & \textit{7} & \textit{0.27} $\pm$ \textit{0.05} \\ 
\textbf{4} & $ 4.08\pm 0.65$ & 5 & $ 0.19\pm0.03 $ \\ 
\textbf{5} & $ 14.05\pm 2.08$ & 17 & $ 0.15\pm0.03 $ \\
\textbf{6} & $ 10.82\pm 1.31$ & 13 & $0.18 \pm 0.04$ \\ 
\textbf{7} & $ 7.37\pm 4.54$ & 13 & $0.23 \pm 0.05$ \\
\textbf{8} & $ 8.62\pm2.18 $ & 11 & $ 0.22\pm 0.04$ \\
\textbf{9} & $ 12.49\pm 1.65$ & 15 & $0.09 \pm 0.03$ \\
\textbf{10} & $ 16.06\pm 3.80$ & 21 & $ 0.17\pm0.04 $ \\
\textbf{11} & $ 18.15\pm 3.20$ & 23 & $ 0.11\pm 0.05$ \\
\textbf{12} & $ 17.34\pm 4.83$ & 23 & $ 0.15\pm0.05 $ \\
\bottomrule
\end{tabular}
\caption{Optimal window size study in En-It ST. (*) For the first three layers, we use standard self-attention.}
\label{tab:windows_it}
\end{table}

\section{Contribution Matrices}
\label{apx:layers_contributions}


Below, we show more examples of contribution matrices for the different languages that have been studied. Note that, although in some cases a diagonal a pattern appears in the first three layers, the diagonality score is still low. Local diagonal patterns are not strictly related to high diagonality, since contributions outside the pattern might be uniformly distributed, and thus difficult to observe in the heatmap. For this reason, contribution matrices can be misleading, and we focus on the use of CL scores to determine which layers should use full attention.

\begin{figure*}[!th]
\centering
\begin{minipage}[b]{\textwidth}
\begin{center}
\includegraphics[width=0.98\textwidth]{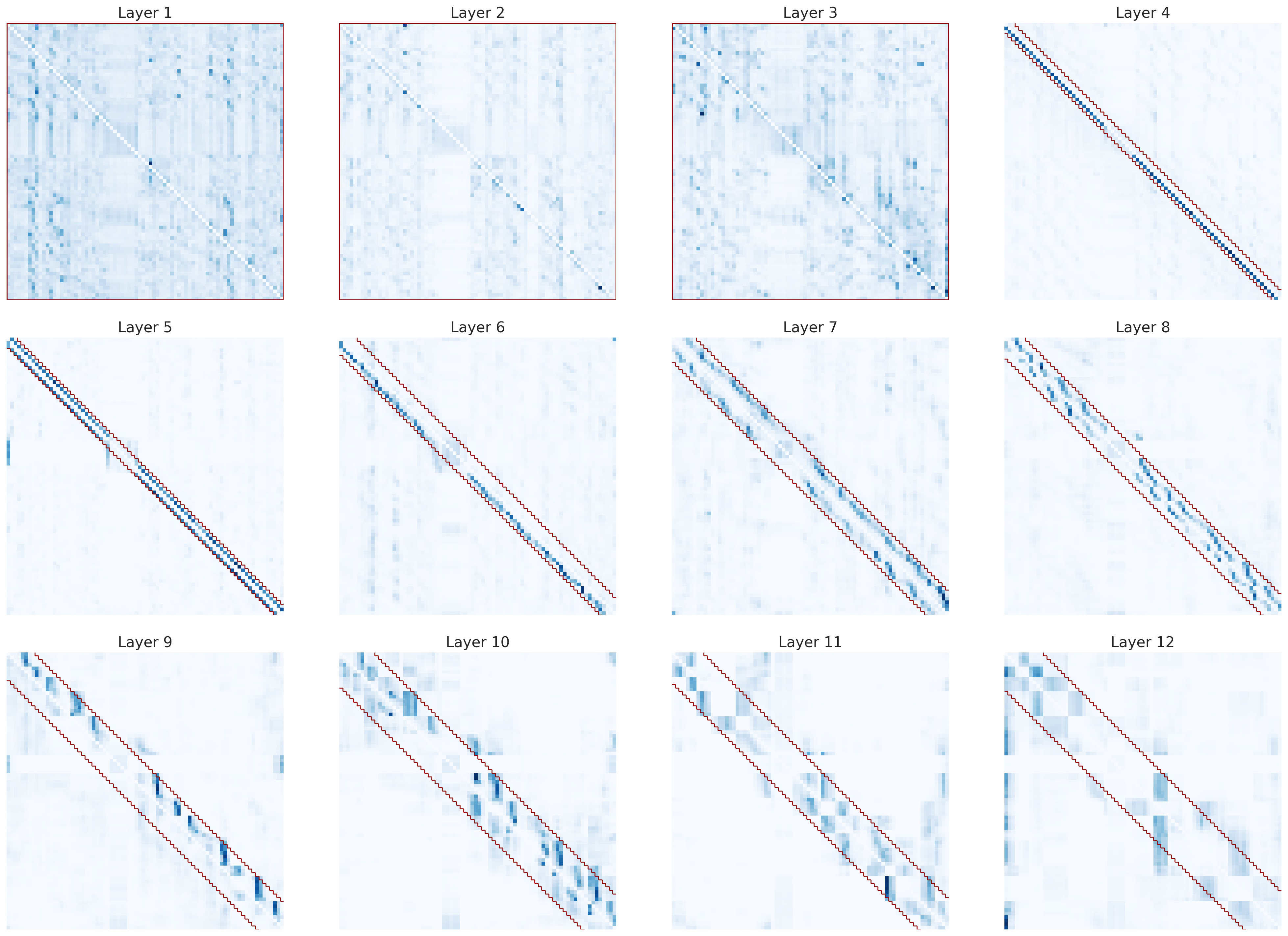}
\end{center}
\end{minipage}\hfill
\caption{Contribution matrices for a sample after En-De ST training.}
\label{fig:layers_contributions_b}
\end{figure*}

\begin{figure*}[!th]
\centering
\begin{minipage}[b]{\textwidth}
\begin{center}
\includegraphics[width=0.98\textwidth]{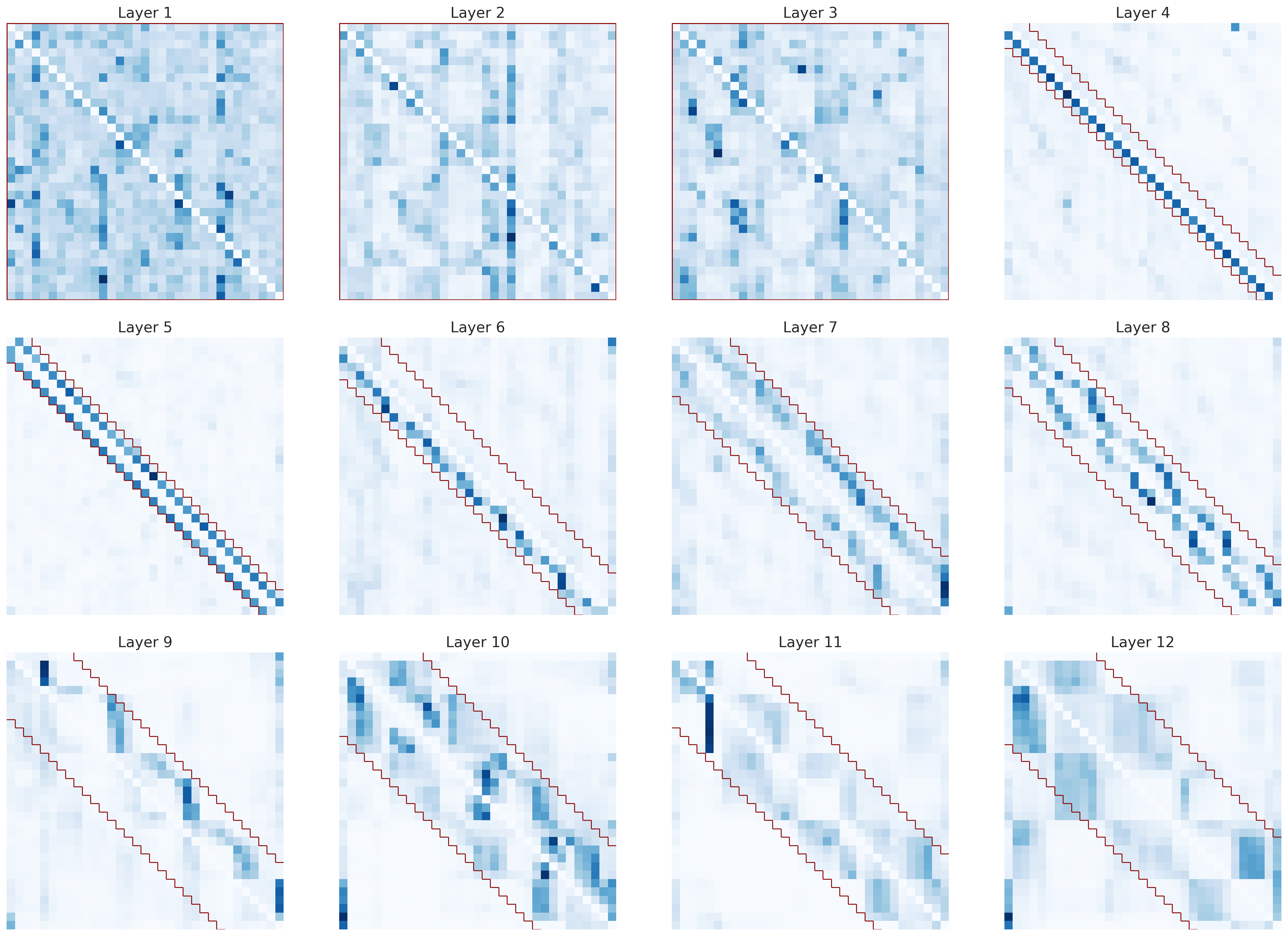}
\end{center}
\end{minipage}\hfill
\caption{Contribution matrices for a sample after En-De ST training.}
\label{fig:layers_contributions_d}
\end{figure*}

\begin{figure*}[!th]
\centering
\begin{minipage}[b]{\textwidth}
\begin{center}
\includegraphics[width=0.98\textwidth]{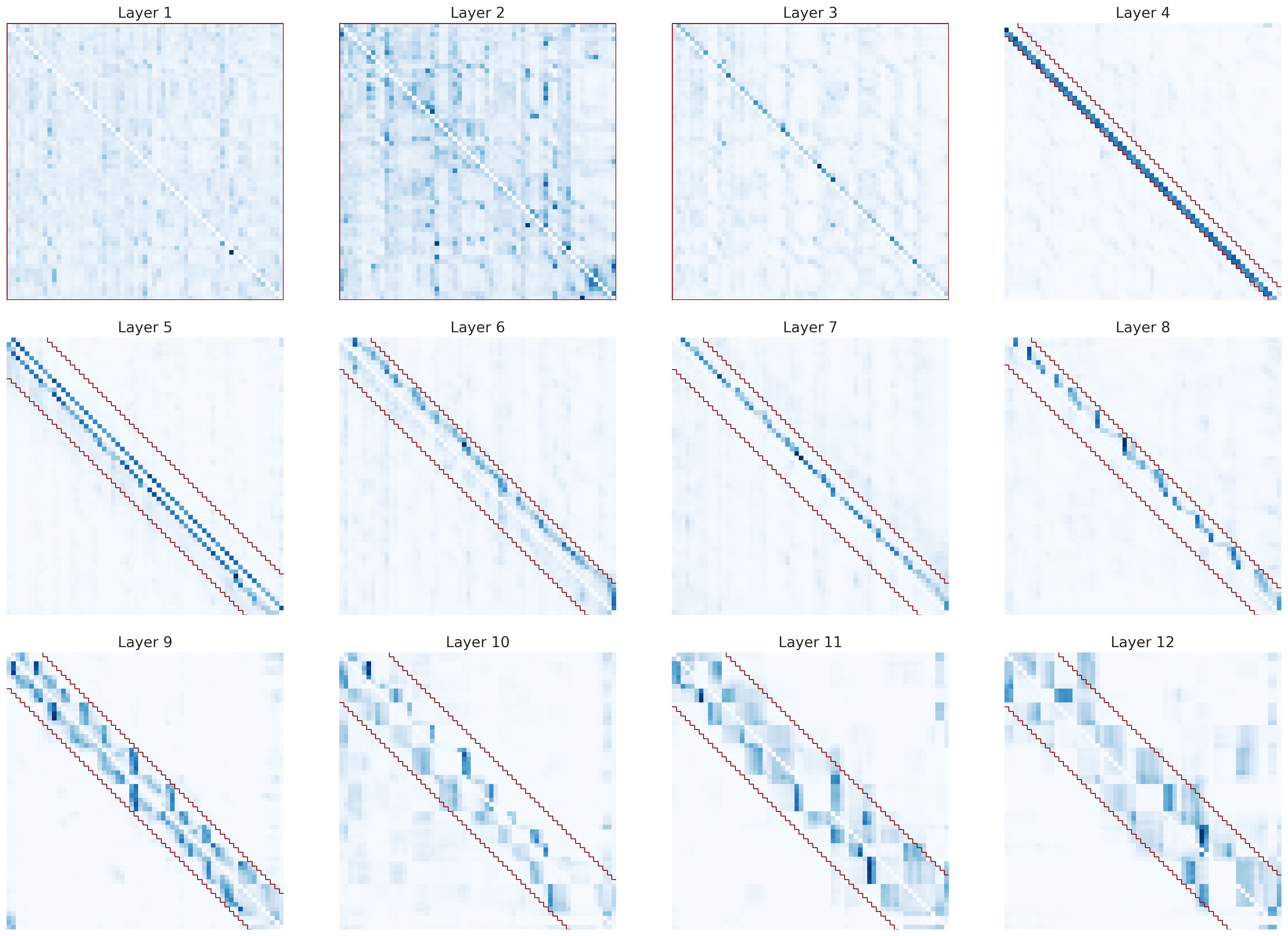}
\end{center}
\end{minipage}\hfill
\caption{Contribution matrices for a sample after En-It ST training.}
\label{fig:layers_contributions_e}
\end{figure*}

\begin{figure*}[!th]
\centering
\begin{minipage}[b]{\textwidth}
\begin{center}
\includegraphics[width=0.98\textwidth]{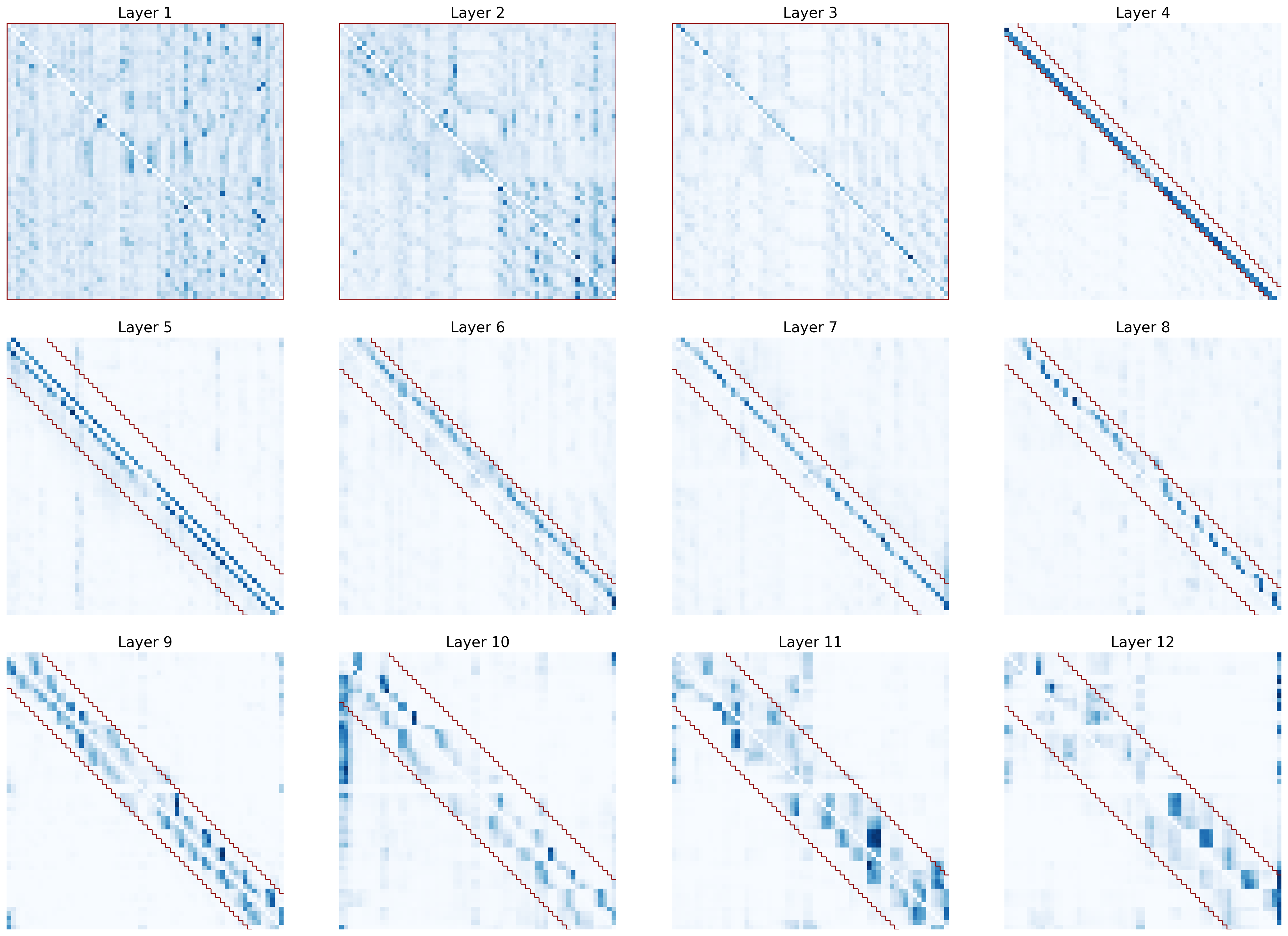}
\end{center}
\end{minipage}\hfill
\caption{Contribution matrices for a sample after En-It ST training.}
\label{fig:layers_contributions_f}
\end{figure*}

\begin{figure*}[!th]
\centering
\begin{minipage}[b]{\textwidth}
\begin{center}
\includegraphics[width=0.98\textwidth]{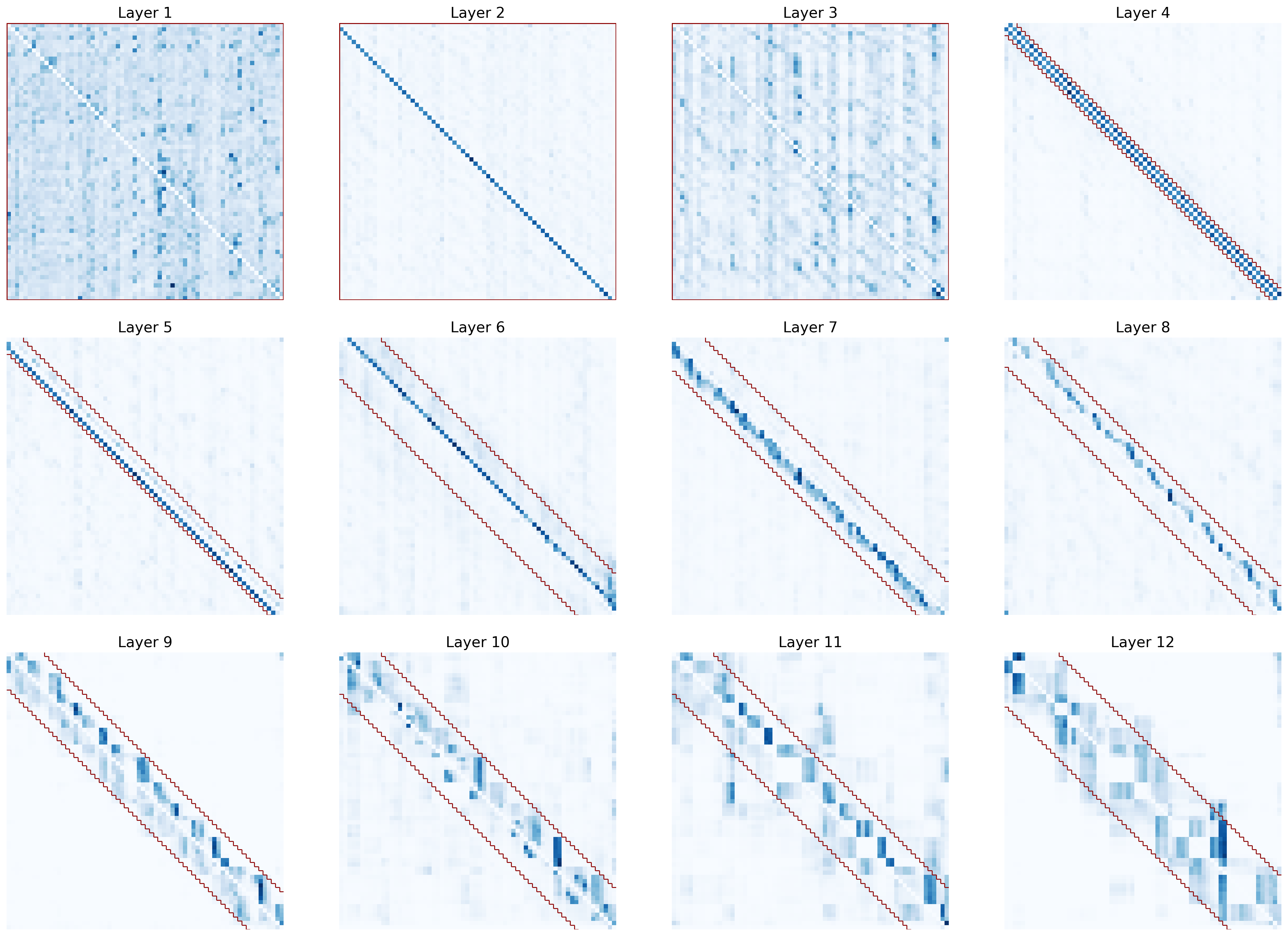}
\end{center}
\end{minipage}\hfill
\caption{Contribution matrices for a sample after En-Es ST training.}
\label{fig:layers_contributions_h}
\end{figure*}

\begin{figure*}[!th]
\centering
\begin{minipage}[b]{\textwidth}
\begin{center}
\includegraphics[width=0.98\textwidth]{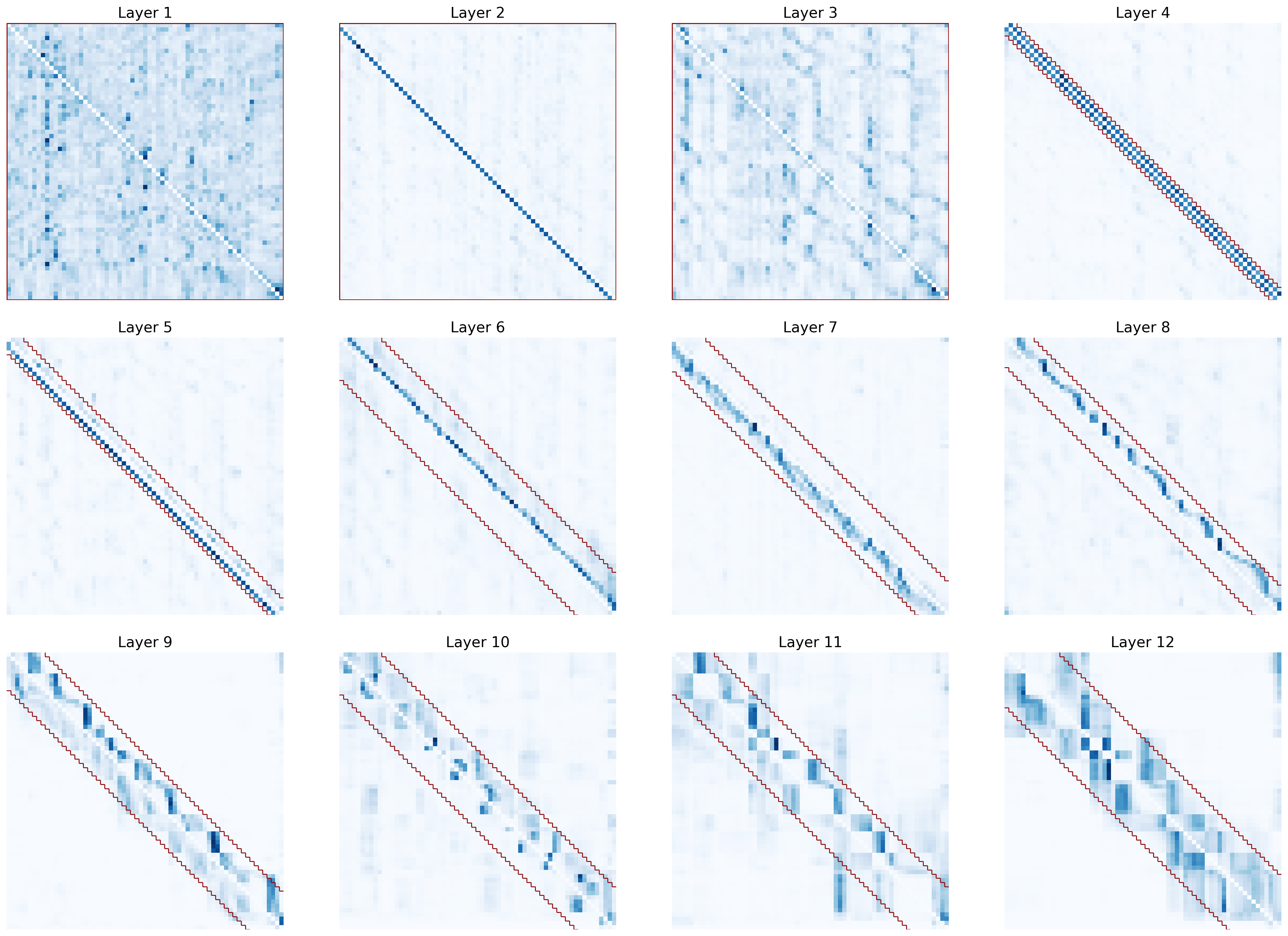}
\end{center}
\end{minipage}\hfill
\caption{Contribution matrices for a sample after En-Es ST training.}
\label{fig:layers_contributions_i}
\end{figure*}

\end{document}